%% file: top.tex
\documentclass[final]{cvpr} 
\usepackage{times}
\usepackage{epsfig}
\usepackage{graphicx}
\usepackage{amsmath}
\usepackage{amssymb}
\usepackage{multirow}
\usepackage{makecell}
\usepackage{placeins}
\usepackage[usenames, dvipsnames]{color} 
\usepackage{subcaption} 

\usepackage[pagebackref=true,breaklinks=true,letterpaper=true,colorlinks,bookmarks=false]{hyperref}
\input{tex/defs}

\def\cvprPaperID{2774} 
\def\confYear{CVPR 2021}

\begin{document}

\title{Deep Active  Surface Models}

\author{Udaranga Wickramasinghe, Pascal Fua\\
Computer Vision Laboratory\\
EPFL\\
{\tt\small \{udaranga.wickramasinghe, pascal.fua\}@epfl.ch}
\and
Graham Knott\\
BioEM Laboratory\\
EPFL\\
{\tt\small  graham.knott@epfl.ch}
}

\maketitle
\thispagestyle{empty}


\input{tex/0_abstract.tex}
\input{tex/1_introduction.tex}
\input{tex/2_related_work.tex}

\input{tex/3_method.tex}

\input{tex/4_experiments.tex}

\input{tex/5_conclussion.tex}

\input{tex/6_acknowledgment.tex}
\input{tex/7_appendix.tex}

{\small
\bibliographystyle{ieee_fullname}
 \bibliography{string,biomed,vision,graphics,learning,optim} 
}

\end{document}

%% file: tex/defs.tex
\newif\ifdraft
\draftfalse

\definecolor{orange}{rgb}{1,0.5,0}
\definecolor{violet}{RGB}{70,0,170}
\definecolor{magenta}{RGB}{170,0,170}
\definecolor{dgreen}{RGB}{0,150,0}

\usepackage[normalem]{ulem}

\ifdraft
\newcommand{\PF}[1]{{\color{red}{\bf PF: #1}}}

\newcommand{\PUW}[1]{{\color{blue}{\bf PUW: #1}}}

\newcommand{\MS}[1]{{\color{dgreen}{\bf MS: #1}}}

\else
\newcommand{\PF}[1]{}

\newcommand{\PUW}[1]{}

\newcommand{\MS}[1]{}

\fi

\newcommand{\comment}[1]{}
\newcommand{\parag}[1]{\vspace{-3mm}\paragraph{#1}}

\DeclareMathOperator{\argmin}{\arg \min}

\newcommand{\bA}{\mathbf{A}}
\newcommand{\bB}{\mathbf{B}}

\newcommand{\bE}{\mathbf{E}}
\newcommand{\bI}{\mathbf{I}}

\newcommand{\bX}{\mathbf{X}}

\newcommand{\mM}{\mathcal{M}}

\newcommand{\mL}{\mathcal{L}}

\newcommand{\mS}{\mathcal{S}}

\newcommand{\unet}[0]{{U-NET}}
\newcommand{\vnet}[0]{{V-NET}}

\newcommand{\voxmesh}[0]{{Voxel2Mesh}}
\newcommand{\pixmesh}[0]{{Pixel2Mesh}}
 
\newcommand{\postprocessdasm}[0]{{PostProc-ASM}} 
\newcommand{\uniformdasm}[0]{{Uniform-DASM}} 
\newcommand{\adaptivedasm}[0]{Ad.-DASM}
\newcommand{\dasm}[0]{{DASM}} 
\newcommand{\meshrcnn}[0]{{Mesh R-CNN}} 

\newcommand{\ternausnet}[0]{{TernausNet}}
\newcommand{\linknet}[0]{{LinkNet34}}
\newcommand{\resnet}[0]{{ResNet50}}
\newcommand{\resnetse}[0]{{ResNet50-SE}}

\usepackage{multirow}

%% file: tex/0_abstract.tex

\begin{abstract}

Active Surface Models have a long history of being useful to model complex 3D surfaces. But only Active Contours have been used in conjunction with deep networks, and then only to produce the data term as well as meta-parameter maps controlling them. In this paper, we advocate a much tighter integration. We introduce layers that implement them that can be integrated seamlessly into  Graph Convolutional Networks to enforce sophisticated smoothness priors at an acceptable computational cost. 

We will show that the resulting Deep Active Surface Models outperform equivalent architectures that use traditional regularization loss terms to impose smoothness priors for 3D surface reconstruction from 2D images and for 3D volume segmentation.

\end{abstract}

%% file: tex/1_introduction.tex

\section{Introduction}

Triangulated meshes are one of the most popular and versatile kind of  3D surface representation. In recent years, one of the most popular approaches to inferring such representations from images has been to use deep networks to produce a volumetric representation and then running a marching-cube style algorithm to create the mesh. However, using marching-cubes tends to introduce artifacts and introduces additional complexities when trying to make the process end-to-end differentiable.  Hence, deep-learning methods that go {\it directly} and without resorting to an intermediate stage from 2D images~\cite{Wang18e,Pan19,Wen19a} and 3D image stacks~\cite{Wickramasinghe20} to 3D triangulated surfaces have recently been proposed. 

Unfortunately, these direct methods are also prone to generating unwanted artifacts such as those shown at the top of Fig.~\ref{fig:teaser}.  State-of-the-art methods handle them by introducing additional regularizing loss terms such as the edge length loss, the normal consistency loss, or the Laplacian loss during training \cite{Gkioxari19}. To be effective without sacrificing reconstruction accuracy, these terms must be carefully weighted, which is typically difficult to achieve. 

\input{fig/teaser}

In this paper, we solve this problem by introducing into the surface generating architecture a special-purpose layer that regularize the meshes using a semi-implicit scheme  that involves recursively solving sparse linear systems of linear equations. It propagates smoothness constraints much faster and more reliably than traditional gradient descent-based energy minimization without requiring much computational power and yields surface meshes that fit the data while remaining smooth, such as those shown at the bottom of Fig.~\ref{fig:teaser}. Furthermore, this scheme enables us to
\begin{itemize}

 \item modulate locally the amount of regularization we impose  so that we  regularize only where it is needed and, hence, preserve accuracy;
 
 \item use meshes consisting of vertices with arbitrary degrees which is not commonly seen in majority of Active Shape Models.
 
 \end{itemize}
Both of these are important to model complex 3D objects that can be smooth in some places and very curvy elsewhere.  

We took our inspiration from the Active Surface Models (ASMs) idea~\cite{Terzopoulos87,Terzopoulos88}, which were first introduced over 30 years ago and also used a semi-implicit optimization scheme to model complex 3D shapes from images. Today, they are mostly used in conjunction with deep networks that are used to compute the data term that is minimized when deforming the models and meta-parameter maps that controls its behavior~\cite{Marcos18,Dong18b}. Even though these methods are end-to-end trainable, they do not embed the ASMs within the contour deforming graph convolution networks as we do. Furthermore, they are limited to 2D contours whereas we handle irregular 3D surface meshes, that is, meshes whose vertices can be of arbitrary degrees. To this end, we propose an original method to compute the derivatives required for back-propagation on such a mesh. 

  
Our contribution therefore is {\it Deep Active Surface Models} (DASMs) that outperform equivalent architectures in which the smoothness constraints are imposed by minimizing a traditional loss function. We will demonstrate this for 3D surface reconstruction from 2D images and for 3D volume segmentation.

%% file: fig/teaser.tex

\begin{figure}[htbp]
\centering
\includegraphics[height=7.0cm]{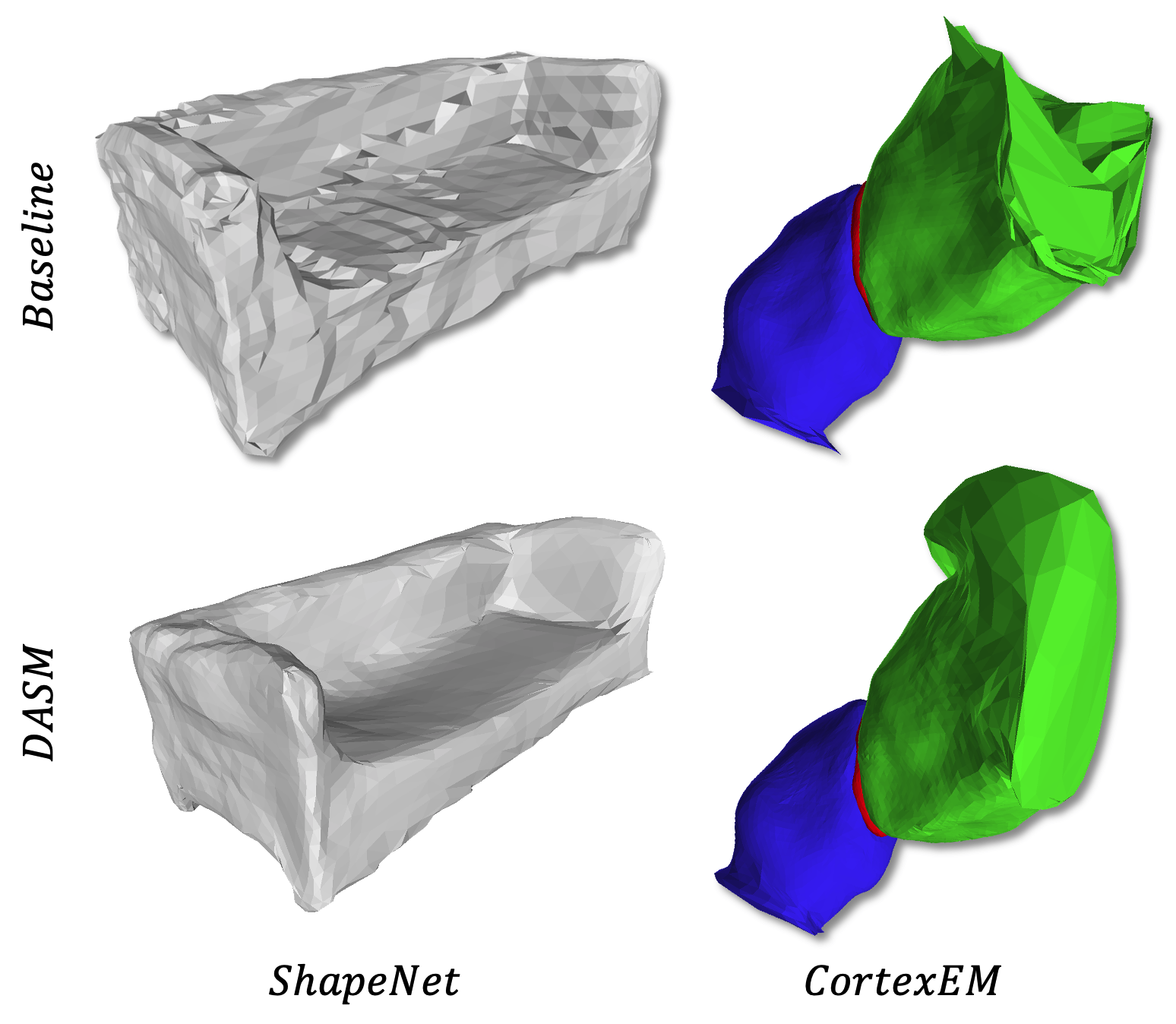}
\caption{\small {\bf Smoothness and Accuracy. } (Top) 3D surface meshes of a couch  modeled from an RGB image and of a synaptic connection segmented from an electron microscopy stack by Mesh R-CNN~\cite{Gkioxari19} and by Vox2Mesh~\cite{Wickramasinghe20}, two state-of-the-art mesh-generating methods. (Bottom) Results using the same backbones augmented by our DASM smoothing layers. The meshes have far fewer artifacts and we will show that they are also more accurate.}
\label{fig:teaser}
\end{figure}

%% file: tex/2_related_work.tex
\UseRawInputEncoding
\section{Related Work} 
\label{sec:related}

\subsection{Active Contour and Surface Models}

Active contour models allow contours to be refined to account for local image properties while preserving global geometric primitives. They were first introduced in~\cite{Kass88}  for interactive delineation and then extended for many different purposes~\cite{Fua96f}. Active surface models operate on the same principle~\cite{Terzopoulos87,Terzopoulos88} but replace the contours by triangulated meshes to model 3D surfaces. They have proved very successful for medical~\cite{McInerney95a,He08a} and cartographic applications~\cite{Fua95c}, among others,  and are still being improved~\cite{Leventon00,Prevost13,Jorstad14a}.

Both active contours and surfaces operate by minimizing an objective function that is a weighted sum of data term derived from the images and a quadratic term that enforces global smoothness. They owe part of their success to the ability to perform the optimization using a semi-implicit scheme that propagates smoothness constraints much faster than gradient descent energy minimization would and gives them superior convergence properties. In our work, we integrate this scheme into our deep architecture. 

\subsection{Deep Surface Models}

Before the advent of deep learning, mesh representations used to be dominant in the field of 3D surface reconstruction~\cite{McInerney95a,Fua96f}. Since then, they have been eclipsed by methods that rely on continuous deep implicit-fields. They represent 3D shapes as level sets of deep networks that map 3D coordinates to a signed distance function~\cite{Park20a,Xu19b} or an occupancy field~\cite{Mescheder19, Chen19c}. This mapping yields a continuous shape representation that is lightweight but not limited in resolution. This representation has been successfully used for single-view reconstruction~\cite{Mescheder19,Chen19c,Xu19b} and 3D shape-completion~\cite{Chibane20}.  


However, for applications requiring explicit surface parameterizations, the non-differentiability of standard approaches to iso-surface extraction, such as the many variants of the Marching Cubes algorithm~\cite{Lorensen87,Newman06}, as well as their tendency to produce artifacts, remain an obstacle to exploiting the advantages of implicit representations. The non-differentiability is addressed in~\cite{Remelli20b} but the artifacts remain. Pixel2Mesh~\cite{Wang18e}  and its newer variants~\cite{Pan19,Wen19a} represent attempts to overcome this difficulty by going {\it directly} from 2D images to 3D surface meshes without resorting to an intermediate stage.  This approach has recently been extended to handle 3D image volumes~\cite{Wickramasinghe20}. These methods rely on graph-convolution layers to iteratively deform an initial mesh to match the target.  While effective, they tend to produce large artifacts that detract both from their accuracy and their usability for further processing. This can be mitigated by introducing regularizing cost terms into the training loss function. However, it is difficult to weigh them properly to remove the artifacts without compromising the accuracy. Our method is designed to address this very issue.

\subsection{Deep Contour Models} 

Similar to Pixel2Mesh and its variants for surface extraction, there exist its 2D counter parts. \cite{Liang20,Peng20} use graph-convolution networks to perform instance segmentation by deforming a contour. The same approach is used in \cite{Ling19} to perform interactive object annotation. 


\subsection{Active Contours/Surfaces and Neural Networks}

Active contour models have been combined with deep networks by exploiting the differentiability  of the active contour algorithm~\cite{Marcos18,Cheng19,Hatamizadeh20}. These approaches use deep networks to produce the data term as well as meta-parameter maps controlling the behavior of the active contour.  This has  also been exploited to correct errors in contour labels used in semantic segmentation~\cite{Acuna19}.  However, we are not aware of any recent work that embeds active surfaces into deep networks. One potential reason is that most current approaches to generating 3D meshes in a deep learning context, such as those discussed above~\cite{Wang18e,Pan19,Wen19a,Wickramasinghe20}, yield irregular meshes in which the vertices can have varying number of neighbors. This makes the computation of the derivatives required for back-propagation non-trivial. This has been addressed in the ASM context by introducing  finite element-based computations that notably complexify the approach~\cite{Lengagne97} or non-easily differentiable elements such as quadric fitting in the neighborhood of some vertices~\cite{Lengagne00}. In this paper, we propose an approach that is back-propagation friendly on potentially large meshes.

%% file: tex/3_method.tex
\section{Method} 
\label{sec:method}

We first introduce the general formulation of Active Surface Models (ASMs) and then show how we can apply it to meshes whose vertices can be of arbitrary degree. Finally, we discuss the integration of ASMs and mesh-deforming Graph-Convolutional Neural Networks (GCNNs), which produces our Deep Active Surface Models (DASMs). 

\subsection{Active Surface Model (ASM)}
\label{sec:asm}

An ASM consists of a surface $\mS(\Phi)$ whose shape is controlled by a vector of parameters $\Phi$ and can be deformed to minimize an objective function $E(\Phi)$, often referred to as an {\it energy}. We first introduce a continuous formulation and then its discretization, which is the one used in practice. 

\parag{Continuous Formulation.} $\mS$ is represented by the mapping from $\mathbb{R}^2$ to  $\mathbb{R}^3$
\begin{equation}\label{eq:surface}
 v: (s,r;\Phi)  \mapsto (v_x(s,r;\Phi), v_y(s,r;\Phi), v_z(s,r;\Phi)) \; ,
\end{equation}
where $(s,r) \in \Omega = [0, 1] \times [0, 1]$. Fig.~\ref{fig:surface} depicts this mapping and its derivatives. $\Phi$ is typically taken to be 

\begin{align}\label{eq:minimize1}
   \Phi^*    &= \argmin_{\Phi} E(\Phi) \; , \\  
    E(\Phi) & = E_{\rm dat} (\Phi) + E_{\rm def} (\Phi) \nonumber ,
\end{align}
\input{fig/surface}
where $E_{\rm dat}$ is a data term that measures how well the surface matches the images and $E_{\rm def}$ is a deformation energy that is smallest when the surface is smooth. $E_{\rm def}$ is often written as
\begin{align}\label{eq:defEnergy}
E_{\rm def} = & \int_{\Omega} w_{10} \left\lVert \frac{\partial v}{\partial s} \right\lVert^2 + w_{01} \left\lVert \frac{\partial v}{\partial r} \right\lVert^2
+ 2w_{11} \left\lVert \frac{\partial^2 v}{\partial s \partial r} \right\lVert^2 \nonumber \\
& + w_{20} \left\lVert \frac{\partial^2 v}{\partial s^2} \right\lVert^2 + w_{02} \left\lVert \frac{\partial^2 v}{\partial r^2} \right\lVert^2 dr ds\; . 
\end{align}

The surface $\Phi^*$ that minimizes the energy $E(\Phi)$satisfies the associated Euler-Lagrange equation \cite{Cohen93,Kass88}
\begin{small}
\begin{align}
F(v) &=
- \frac{\partial}{\partial s} \Bigg ( w_{10}  \frac{\partial v}{\partial s} \Bigg )
- \frac{\partial}{\partial r} \Bigg ( w_{01}  \frac{\partial v}{\partial r} \Bigg ) \label{eq:euler} \\
\!+ 2 & \frac{\partial ^2}{\partial s \partial r} \Bigg ( w_{11}  \frac{\partial^2 v}{\partial s \partial r}\Bigg )
\!+ \! \frac{\partial}{\partial^2 s} \Bigg ( w_{20}  \frac{\partial^2 v}{\partial s^2} \Bigg )
\!+ \! \frac{\partial}{\partial^2 r} \Bigg ( w_{02}  \frac{\partial^2 v}{\partial r^2} \Bigg ) , \nonumber
\end{align} 
\end{small}
where $F=-\nabla E_{\rm dat}$.

\parag{Discrete Formulation.}

When $\mS(\Phi)$ is discretized and represented by a triangulated mesh $\mM(\Phi)$, $\Phi$ becomes the $3N$-vector built by concatenating the 3D coordinates of its $N$ vertices. Using the finite-difference approximation described in the appendix, Eq~\ref{eq:euler} can be written in matrix form as
\begin{equation}\label{eq:minimize2} 
\bA \Phi^*  =   F(\Phi^*) \; ,
\end{equation}
where $F$ is the negative gradient of $E_{\rm dat}$ with respect to $\Phi$. Because $\bA$ is usually non-invertible, given an initial value $\Phi^0$, a solution to this equation can be found by iteratively solving 
\begin{align}
 \alpha ( \Phi^t - \Phi^{t-1} ) + \bA \Phi^t & = F(\Phi^{t-1})   \; , \nonumber  \\
    \Rightarrow  (\bA + \alpha \bI) \Phi^t &=  \alpha  \Phi^{t-1} + F(\Phi^{t-1}) \; , \label{eq:viscosity}
\end{align}
where $\bI$ is the identity matrix. When the process stabilizes, $\Phi^{t} = \Phi^{t-1}$ and is a solution of Eq.~\ref{eq:minimize2}.

The strength of this semi-implicit optimization scheme is that it propagates smoothness constraints much faster than traditional gradient descent that minimizes  energy $E(\Phi)$ and at a low computational cost because $\bA$ is sparse, which means that the linear system of Eq.~\ref{eq:viscosity} can be solved efficiently. In this scheme $\alpha$ plays the role of the inverse of a step size: When $\alpha$ is large enough for the Froebinius norm of $\alpha \bI$ to be much larger than that of $\bA$, the optimizer performs a steepest gradient step given by $F(\Phi^{t-1})$ with learning rate $\frac{1}{\alpha}$ at each iteration. Conversely, when $\alpha$ is small, $\bA$ dominates and much larger steps can be taken.

In the original deformable contour models~\cite{Kass88},  the matrix $\bA + \alpha \bI$ was never inverted. Instead Eq.~\ref{eq:viscosity} was solved by LU decomposition.  Instead, to implement this effectively on a GPU using sparse tensors and to speed up the computations of the losses and their derivatives, we approximate  the inverse of $(\bA + \alpha \bI)$ using the Neumann series
\begin{align}\label{eq:inverse} 
& (\bA + \alpha \bI) ^{-1} \approx \sum_{n=0}^{K} (-1)^n \left(\frac{1}{\alpha}\right)^{n+1}A^n \; .
\end{align} 
and use it to solve Eq.~\ref{eq:viscosity}.  We use $K=4$, which yields a sufficiently good approximation of  actually solving Eq.~\ref{eq:viscosity}. 

\parag{Computing the Regularization Matrix  $\bA$.}

In most traditional ASMs, the meshes are either square or hexagonal and regular, which makes the computation of the derivatives of the mesh vertices possible using finite-differences and, hence, the regularization matrix $\bA$ of Eq.~\ref{eq:minimize2} easy to populate. 

When the mesh is triangular and irregular, vertices can have any number of neighbors and the computation becomes more complex. Nevertheless the required derivatives, of order 2 and 4, can still be expressed as finite differences of weighted sums of vertex coordinates where the weights are barycentric coordinates of small perturbations of the original vertices. This is explained in more details in the appendix.

\subsection{Deep Active Surface Model (DASM)}
\label{sec:dasm}

The update equation in a typical mesh-deforming graph-convolutional neural network (GCNN) that plays the same role as that of Eq.~\ref{eq:viscosity} is
\begin{equation}
\Phi^t =  \Phi^{t-1} +  \frac{1}{\alpha} F(\Phi^{t-1},\bX^{t-1}) \; ,
\label{eq:grad}
\end{equation}
where $F$ denotes the negative gradient of the loss function calculated using the feature vector $\bX^{t-1}$ associated with the mesh parameters $\Phi^{t-1}$. In the case of our deep active surface models, it becomes
\begin{equation}
(\bA + \alpha \bI) \Phi^t =  \alpha  \Phi^{t-1} +  F(\Phi^{t-1},\bX^{t-1}) \; ,
\label{eq:impl}
\end{equation}
as in Eq.~\ref{eq:viscosity}. In Eq.~\ref{eq:grad}, the loss function typically includes a regularization term to keep the mesh smooth, whereas in Eq.~\ref{eq:impl} our semi-implicit scheme enforces smoothness by solving the linear equation. 


\parag{Uniform vs Adaptive DASMs}

Eq.~\ref{eq:impl} forms the basis of the simplest version of our DASMs, which we will refer to as Uniform DASMs because the same amount of smoothing is applied across the whole mesh. This may result in under- or over-smoothing because some parts of the objects require more smoothing while some parts do not. 
 
To account for this, we also introduce Adaptive DASMs that are designed to smooth only where necessary, as indicated by an auxiliary metric. Experimentally, adaptive smoothing is required when the GCNN produces particularly large deformations but only in a very specific part of the mesh or fails to smooth-out artifacts produced by mesh initialization algorithms. This could be eliminated by strongly smoothing everywhere but would degrade accuracy in high-curvature areas. 

%

To solve this problem, we begin by using the approximation of $(\bA + \alpha \bI)^{-1}$ from Eq.~\ref{eq:inverse} to rewrite the evolution equation of Eq.~\ref{eq:impl} as
\begin{align}
\Gamma^t & =  \Phi^{t-1} + \frac{1}{\alpha} F(\Phi^{t-1},\bX^{t-1}) \; , \nonumber \\
\Phi^t        & = (\bI +  \sum_{n=1}^{K} (-1)^n \left(\frac{1}{\alpha}\right)^{n} \bA^n)  \Gamma^t \; , \label{eq:smoothUpdate1} \\
& = \Gamma^t + \bB \Gamma^t \mbox{ with } \; \bB = \sum_{n=1}^{K} (-1)^n \left(\frac{1}{\alpha}\right)^{n} \bA^n \; . \nonumber 
\end{align}
$\Gamma^t$ represents  $\Phi^{t-1}$ incremented by the negative gradient of the loss function $F(\Phi^{t-1})$ but not yet smoothed. In other words, we have rewritten the smoothing operation that transforms $\Gamma^t$ into $\Phi^t$ as simply adding $\bB\Gamma^t$ to $\Gamma^t$ . This gives us the freedom to decide where we want to smooth and where we do not by introducing a diagonal matrix $\Lambda$ and rewriting the update rule of Eq.~\ref{eq:smoothUpdate1} as 
\begin{align}
\label{eq:smooth_update}
 \Phi^t       & = \Gamma^t + \Lambda \bB \Gamma^t \; .
\end{align}
This update rule is similar to the one of the Adagrad algorithm \cite{Duchi11}. Here, each diagonal component $\lambda_{i,i}$ of $\Lambda$ rescales the corresponding component  $(\bB \Gamma)_i$ of $\bB \Gamma$. In Adagrad, adaptive re-scaling is a function of past gradients. Here we take it to be a function of current surface gradients because we have observed that $|\bB \Gamma^t|_{i}$ tends to grow large when the facets increase in size and smoothing is required, and remains small otherwise. We therefore take the diagonal values of $\Lambda$ to be
\begin{align}
\lambda_{ii} = \sigma (|\bB \Gamma^t|_{i}; \beta, \gamma) \; ,
\end{align} 
where $\sigma$ is the Sigmoid function and $\beta, \gamma$ are its steepness and midpoint. In this way, for small values of $|\bB \Gamma^t|_{i}$, there is almost no smoothing, but for larger ones there is. Fig.~\ref{fig:adaptive_dasm_demo} illustrates this behavior. 

\input{fig/adaptive_dasm_demo}
\parag{Recursive smoothing}

\input{fig/asm_demo}

Any single DASM step given by Eq.~\ref{eq:smooth_update} can only rectify a finite amount of deformations. To mitigate this, we perform more than one adaptive-smoothing step in-between gradient updates. During these additional smoothing steps no gradient update is done and we use $F(\Phi^{t-1},\bX^{t-1}) = 0$. In practice we perform these steps until $\|\Phi^t - \Phi^{t-1} \|< \epsilon$, where $\epsilon$ is a preset constant. Fig.~\ref{fig:asm_demo} illustrates this process. 


\parag{Loss terms}
\label{sec:loss_terms}
In architectures such as Mesh R-CNN~\cite{Gkioxari19} and Voxel2Mesh~\cite{Wickramasinghe20}, a loss term is used to supervise the output of each mesh-refinement stage. We follow the same approach and add a loss term at the end of each DASM module. We write it as
\begin{align}
\mL & = \mL_{data} + \lambda \mL_{reg.}\; , \label{eq:totalLoss} \\ 
\mL_{data}  &= \lambda_{cf.} \mL_{cf.} + \lambda_{n. dist.} \mL_{n. dist.}\;,  \nonumber  \\ 
\mL_{reg.}  &= \lambda_{edge} \mL_{edge} + \lambda_{Lap.} \mL_{Lap.} + \lambda_{n. cons.} \mL_{n. cons.}\; . \nonumber  
\end{align} 
Here $\mL_{Cf.} , \mL_{n. dist.} $ are Chamfer and Normal distances \cite{Gkioxari19} and $\mL_{edge}, \mL_{Lap.}, \mL_{n. cons.} $  are edge length loss, Laplacian loss and normal consistency loss, respectively \cite{Wang18e}. All these loss terms are used in Voxel2Mesh~\cite{Wickramasinghe20} except $\mL_{Norm.}$. Similarly, they are all used in Mesh R-CNN~\cite{Gkioxari19} except $ \mL_{Lap.} $ and $\mL_{Norm.}$.

%% file: fig/surface.tex

\begin{figure}[htbp]
\centering

\includegraphics[height=5.0cm]{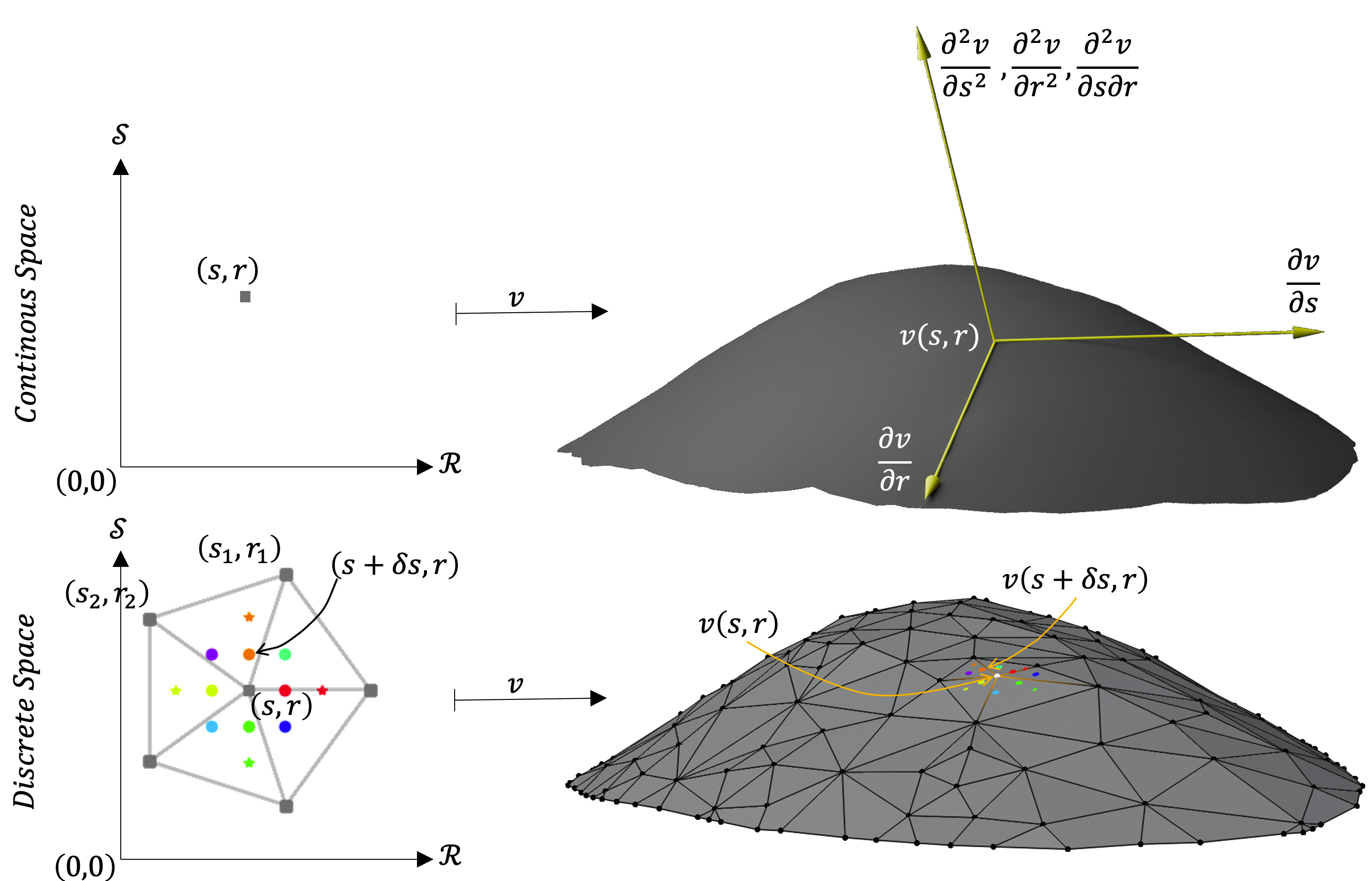}
\caption{\small {\bf Derivatives.} {\bf Top.} The surface is represented by a differentiable mapping $v$ from $\mathbb{R}^2$ to $\mathbb{R}^3$. {\bf Bottom.} After discretization, let $v(s,r)$ be a vertex. To approximate derivatives with respect to $s$ using finite differences, we need to estimate quantities such as $v(s+\delta s,r)$ where $\delta s$ is small. To this end, we estimate the barycentric coordinates $\lambda$, $\lambda_1$, and $\lambda_2$ of $v(s+\delta s,r)$  in the facet it belongs to and take $v(s+\delta s,r)$ to be $\lambda v(s,r) + \lambda_1 v(s_1,r_1)+\lambda_2 v(s_2+r_2)$, where $v(s_1,r_1)$ and $v(s_2,r_2)$ are the other two vertices of the facet. The same operation can be performed for derivatives with respect to $r$.}
\label{fig:surface}
\end{figure}

%% file: fig/adaptive_dasm_demo.tex

	\begin{figure}
		\centering
		
		\begin{subfigure}[b]{0.5\textwidth}
			\centering
	\includegraphics[height=3.75cm]{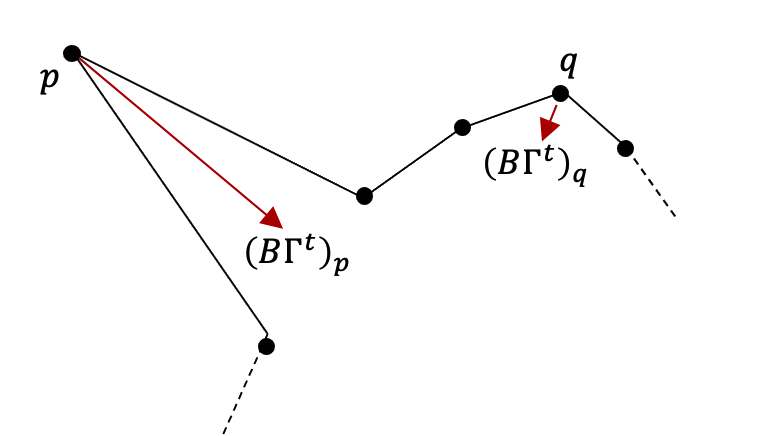}\\
	\vspace{-2.5mm}
	\caption{Starting mesh $\Gamma^{t}$}
			\label{fig:Ng1} 
		\end{subfigure}
		\begin{subfigure}[b]{0.5\textwidth}
			\centering
	\includegraphics[height=3.0cm]{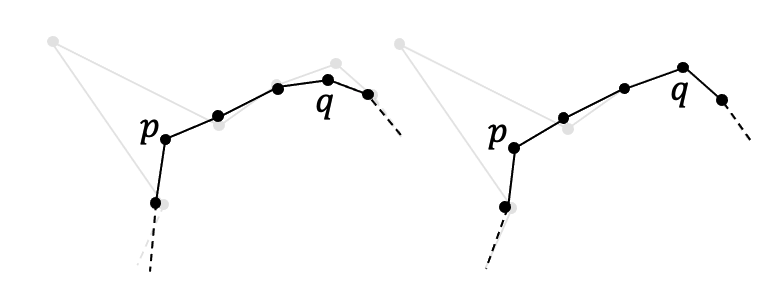}\\
		\vspace{-4mm}
\caption{ Uniform DASM   \hspace{2mm} (c) Adaptive DASM}
			\label{fig:Ng2}
		\end{subfigure}
		\vspace{-6mm}
		\caption{\small{\bf Uniform vs Adaptative Smoothing.} (a) Mesh at time $t$. In general, $|\bB \Gamma^t|_q < |\bB \Gamma^t|_p$. (b) When using enough uniform smoothing to remove the irregularity  at point $p$, the mesh will typically be oversmoothed at $q$. (c) When using adaptative smoothing, the mesh is smoothed around $p$ but not oversmoothed around $q$.}
		\label{fig:adaptive_dasm_demo}
	\end{figure} 

%% file: fig/asm_demo.tex

\begin{figure*}
\centering
\includegraphics[height=2.5cm]{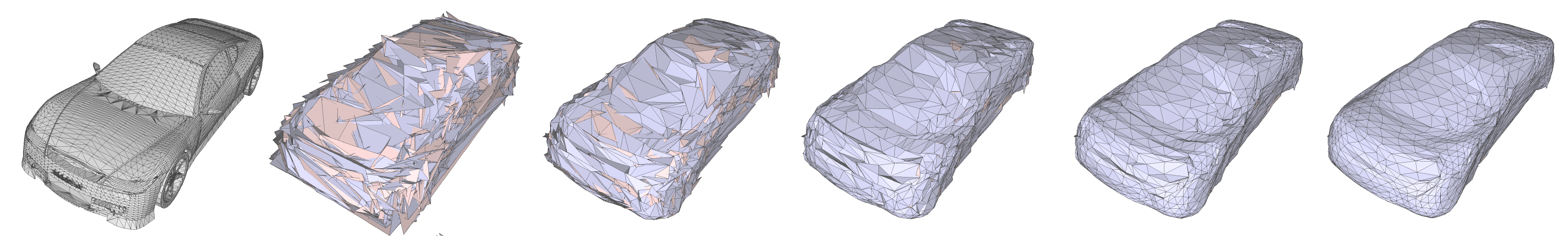}\\
(a) Ground truth\hspace{0.5cm} (b) Input \hspace{0.5cm}(c) ASM 1-step \hspace{0.5cm}(d) ASM 3-steps\hspace{0.5cm}(e) ASM 5-steps \hspace{0.5cm}(f) ASM 7-steps
\vspace{-3mm}
\caption{\small {\bf Performing multiple smoothing steps} We start with a very noisy mesh produced by the algorithm of~\cite{Gkioxari19} run without regularization. We have colored front-faces and back-faces of the meshes in blue and pink respectively for better visualization. After running one step of surface evolution given by Eq.~\ref{eq:viscosity}, we obtain the mesh of Fig.~\ref{fig:asm_demo} (c). Subsequent meshes shown in Fig.~\ref{fig:asm_demo} (d,e,f) are obtained by continuing the surface evolution for 3, 5, and 7 steps respectively. In these subsequent steps we set $F(\Phi^{t-1})$  to zero. 
}
\label{fig:asm_demo}
\end{figure*}

%% file: tex/4_experiments.tex
\section{Experiments} 
\label{sec:experiments}
In this section, we test DASM's ability to predict 3D surfaces from 2D images on Shapenet~\cite{Chang15}  and to extract 3D surfaces from electron microscopy image stacks.
 
\subsection{From 2D Images to 3D Surfaces}
\label{sec:2D3D}

\noindent For prediction of 3D surfaces from 2D images, we benchmark our Adaptative DASM, which we will refer to as \adaptivedasm{}, on the ShapeNet dataset~\cite{Chang15}. 

\parag{Baselines.}

We use \meshrcnn{}~\cite{Gkioxari19} both as a baseline and as the backbone of our network because, among methods that use explicit surface representations, it is currently reported as yielding the best results on ShapeNet. We also compare against \pixmesh{}~\cite{Wang18e}. 

\parag{Dataset.} ShapeNet is a collection of 3D textured CAD models split into semantic categories. As in the  \meshrcnn{} experiments, we use ShapeNetCore.v1 and corresponding rendered images from~\cite{Choy16}. They are of size $137 \times 137$ and have been captured from 24 random viewpoints. We use the train / test splits of \cite{Gkioxari19}, that is, 35,011 models seen in 840,189 images for training and 8,757 models seen in 210,051 images for testing. We use 5\% of the training data for validation purposes.

\parag{Metrics.} We use the same metrics as in \meshrcnn{}. They are the Chamfer distance, Normal distance, and $F1^{\tau}$ at $\tau = 0.1, 0.3$ and $0.5$. For the Chamfer distance a lower value is better while a higher value is better for the others.  

\parag{Implementation.} 

We use the publicly available Pytorch implementation of \meshrcnn{} and incorporate Adaptative DASM layers after each mesh-refinement stage. We also add a Uniform DASM layer after the cubify operation to make the input to mesh refinement stages smooth. We train the networks for 12 epochs using Adam optimizer~\cite{Kingma14a} with a learning rate $10^{-4}$. We set $\alpha = 1, \beta = 6000$ and $\gamma = 15$. 

\meshrcnn{} only uses the $\mL_{edge}$ term of Eq.~\ref{eq:totalLoss} for regularization purposes when training on ShapeNet and turns off the term $\mL_{Lap.}$ because, according to remarks on Github by the authors,  it has not helped to improve the results. For a fair comparison, we therefore do the same.  In this setup, $\mL_{edge}$, which penalizes increases in edge-length, is the only other source of geometric regularization besides the one we provide with our DASM layers. We will therefore experiment with different values of $\lambda_{edge}$, the weight parameter in Eq.~\ref{eq:totalLoss}  that controls how much influence it is given. 

\parag{Results.}

\input{results/shapenet_baselines}
\input{results/shapenet_diff_lambda.tex}
\input{results/shapenet_ablation}

\input{fig/results_shapenet}
\input{fig/results_shapenet_lambda}

We provide qualitative results in Fig.~\ref{fig:shapenet_results} and report quantitative results in Table~\ref{table:shapenet_results} for $\lambda_{edge}=0.2$.  \adaptivedasm{} outperforms  \pixmesh{} and boosts the performance of \meshrcnn{}. Furthermore, the meshes it produces are of much a better visual quality.

In Table~\ref{table:shapenet_results_diff_lambda}, we report similar results for different values of $\lambda_{edge}$, which are depicted qualitatively by Fig.~\ref{fig:shapenet_results_lambda}. The trend is the same for  $\lambda_{edge}=0.6$ and  $1.0$. However, for $\lambda_{edge}=0.0$, the Chamfer distance for \meshrcnn{} is lowest even though the resulting meshes are extremely noisy, as can be seen in the leftmost column of Fig.~\ref{fig:shapenet_results_lambda}. Our interpretation for this somewhat surprising result is that, in this regime, the meshes produced by \meshrcnn{} are so noisy that DASM smoothing takes them away from the data they are trying to fit and degrades the Chamfer distance. In any event, even though the Chamfer distance is low, this can hardly be considered as a good results, hence confirming the observation made in~\cite{Gkioxari19,Wang18e} that this metric might not be the best to evaluate the quality of a mesh. 
When $\lambda_{edge}=1.0$, the difference between \dasm{} and \meshrcnn{} performance is not statistically significant and this is because, when using higher $\lambda_{edge}$, there are not many anomalies for the \dasm{} to fix.

\parag{Ablation Study}

It could be argued that we would have gotten similar results by simply smoothing our meshes as a post-processing step. To demonstrate this is not the case, we implemented \postprocessdasm{} that starts with \meshrcnn{} model trained with $\lambda_{edge}=0.2$ that is then adaptively smoothed by running several times the surface evolution update of Eq.~\ref{eq:smooth_update}. In Table.~\ref{table:shapenet_ablation}, we compare \postprocessdasm{} against \adaptivedasm{} and the results are clearly worse. We also compare \adaptivedasm{} against \uniformdasm{}, which clearly shows the benefit of the adaptive scheme of Section~\ref{sec:dasm}. 




\parag{Failure modes} 

The main source of failures are anomalies produced by the GCNN that are too large to be rectified.  Remaining ones are denoted by blue arrows in Fig~\ref{fig:shapenet_results}.

 
\subsection{From 3D Image stacks to 3D Surfaces}

\input{fig/results_synapses}

In this section we benchmark our approach on CortexEM dataset and compare DASM against  \voxmesh{}~\cite{Wickramasinghe20} and several other baselines. 

\parag{Baselines.} 

We use \voxmesh{} both as a baseline and as the backbone of our network. We also compare our performance against several architectures popular in the biomedical imaging community~\cite{Cicek16,Milletari16,Iglovikov18,Shvets18,Kavur19b}. 

\parag{Dataset.} 

CortexEM is a $500\times 500 \times 200$ FIB-SEM image stack of a mouse cortex. From this 26 sub-volumes with dimension $96 \times 96 \times 96$ were extracted so that each one contains a synaptic junction that is roughly centered. 14 sub-volumes in the first 100 slices in the image stack are used for training and the remaining 12 in the next 100 slice are used for testing. The task is to segment the pre-synaptic region,
post-synaptic region, and synaptic cleft as shown in Fig.~\ref{fig:cortex_results}.

\parag{Metrics.} 

As in~\cite{Wickramasinghe20} and many other papers, we use the intersection-over-union (IoU) as a measure of quality for volumetric segmentation. To compare the meshes, we use the Chamfer distance as in~\cite{Wickramasinghe20}. We repeat our experiments 3 times for each model and report the mean and the standard deviation.
\input{results/crotex}

\input{results/crotex_chamfer}

\parag{Implementation.} 

As we did with \meshrcnn{} in Section~\ref{sec:2D3D}, we incorporate \adaptivedasm{} layers into a \voxmesh{} backbone. We train the networks for 150000 iterations using Adam optimizer~\cite{Kingma14a} with a learning rate of $10^{-4}$. We set $\alpha = 1, \beta = 6000$ and $\gamma = 45$. 

To match the conditions in~\cite{Wickramasinghe20}, \adaptivedasm{} was trained using $\lambda_{edge} = \lambda_{Lap.} = \lambda_{n. cons.} = 0.25$ in Eq.~\ref{eq:totalLoss}. For comparison purposes, we also use $\lambda_{edge} = \lambda_{Lap.} = \lambda_{n. cons.} = 0.025$. To differentiate two versions, we add the regularization coefficients as a subscript to model names. This gives us \voxmesh{}$_{0.25}$, \adaptivedasm{}$_{0.25}$, \voxmesh{}$_{0.025},$~and~ \adaptivedasm{}$_{0.025}$.

\parag{Results}

We report quantitative results in Table~\ref{table:main_results_cortex} in IoU terms and in Table~\ref{table:main_results_chamfer} in Chamfer distance terms. Fig.~\ref{fig:cortex_results} depicts qualitative results for  \adaptivedasm{}$_{0.25}$, and  \voxmesh{}$_{0.025}$. \adaptivedasm{}$_{0.025}$ easily outperforms \voxmesh{} when segmenting pre and post synaptic regions.  For the smaller synaptic junction,  \voxmesh{} and \dasm{} are statistically equivalent  because, unlike for the other two regions, their shapes are simple and there is not much  improvement for DASM to make. In fact, the best result is obtained by  a vanilla U-Net.

 \input{results/cortex_timing}

\parag{Computation time} We report average execution time for a single forward and backward pass for \voxmesh{} and \dasm{} in Table~\ref{table:running_time}. We run this test on a single Tesla V100 GPU. We have implemented the ASM module using custom CUDA kernels and uses sparse tensors. Assembling the regularization matrix and performing the update of Eq.~\ref{eq:smooth_update} adds a 40\% overhead, which is reasonable giving how large the matrices we deal with are.

%% file: results/shapenet_baselines.tex

\begin{table}[!htbp]

\begin{small}
	\begin{tabular}{l|ccccc}
		\hline
		  		~					& Chf. ($\downarrow$)& Normal &  $F^{0.1}$  &  $F^{0.3}$  &  $F^{0.5}$\\ \Xhline{3\arrayrulewidth}
	  	 \pixmesh{}  &         0.241                & 0.701  &   31.6    &   77.3 &91.3   \\
		 \meshrcnn{}  &   0.189      			  &        0.691                      &     32.8   & 80.4 &  \textbf{{92.6}}\\
		 \adaptivedasm{} \hspace{-1mm}		 &   \textbf{0.183}      &       \textbf{0.727}		&    \textbf{33.5}  & \textbf{81.0} &      \textbf{{92.6 }}          \\ \hline
	\end{tabular}
\end{small}
\vspace{-2mm}
\caption{\small {\bf Comparative results on ShapeNet.  }} 
\label{table:shapenet_results}

\end{table}

%% file: results/shapenet_diff_lambda.tex
\newcommand\x{2.0}
\newcommand\y{1.0}
\begin{table}[!htbp]

\begin{small}
	\begin{tabular}{l|l|ccccc}
		\hline
	$\lambda_{e.}$	 			&  				\hspace{-\x mm}		& \hspace{-\x mm} Chf. ($\downarrow$)  	& \hspace{-\x mm}Norm. 				& \hspace{-\x mm} $F^{0.1}$ 		 	 & \hspace{-\x mm} $F^{0.3}$  			& \hspace{-\x mm} $F^{0.5}$\\ \Xhline{3\arrayrulewidth}
	\multirow{2}{*}{1.0}    	& \hspace{-\y mm}\meshrcnn{}  	    & \hspace{-\x mm}	0.232             & \hspace{-\x mm}    0.684    		& \hspace{-\x mm}\textbf{29.7}  	 	 & \hspace{-\x mm} \textbf{76.6}  		& \hspace{-\x mm}   \textbf{89.4}  \\
								& \hspace{-\y mm}\adaptivedasm{}	& \hspace{-\x mm}\textbf{0.231}    & \hspace{-\x mm} \textbf{0.691}  	& \hspace{-\x mm}29.3      	 	 & \hspace{-\x mm}    \textbf{76.6}   	& \hspace{-\x mm}{89.3}       \\ \hline
 
	\multirow{2}{*}{0.6}    	& \hspace{-\y mm}\meshrcnn{}        & \hspace{-\x mm}   0.212     			& \hspace{-\x mm} 0.681             &  \hspace{-\x mm}30.6 			 	 & \hspace{-\x mm}79.2 					& \hspace{-\x mm}  \textbf{91.4}\\
								& \hspace{-\y mm}\adaptivedasm{} 	& \hspace{-\x mm}   \textbf{0.206}      & \hspace{-\x mm}   \textbf{0.693}	&  \hspace{-\x mm}\textbf{31.4}       & \hspace{-\x mm}\textbf{79.5}			& \hspace{-\x mm}     \textbf{91.4}         \\ \hline

	\multirow{2}{*}{0.2}    	& \hspace{-\y mm}\meshrcnn{} 	    & \hspace{-\x mm}   0.189      		    & \hspace{-\x mm}       0.691           &  \hspace{-\x mm}32.8   		 & \hspace{-\x mm} 80.4 				& \hspace{-\x mm}  \textbf{92.6}\\
								& \hspace{-\y mm}\adaptivedasm{} 	& \hspace{-\x mm}   \textbf{0.183}      & \hspace{-\x mm}      \textbf{0.727}  &  \hspace{-\x mm}\textbf{33.5}  & \hspace{-\x mm} \textbf{81.0} 		&  \hspace{-\x mm}    \textbf{{92.6}}         \\ \hline
	 
	\multirow{2}{*}{0.0}    	& \hspace{-\y mm}\meshrcnn{} 	    & \hspace{-\x mm}   \textbf{0.144}      & \hspace{-\x mm}       0.713           &   \hspace{-\x mm}\textbf{35.8}  & \hspace{-\x mm} \textbf{85.1}		&  \hspace{-\x mm}  \textbf{94.2} \\
								& \hspace{-\y mm}\adaptivedasm{} 	& \hspace{-\x mm}    0.167              & \hspace{-\x mm}       \textbf{0.718} 	&     \hspace{-\x mm}34.3     	  	 &\hspace{-\x mm}  84.3 				&  \hspace{-\x mm}       93.9          \\ \hline
	\end{tabular}
\vspace{-2mm}
\caption{\small {\bf Results on ShapeNet} as a function of $\lambda_{edge}$.}
\label{table:shapenet_results_diff_lambda}

\end{small}
\end{table}

%
%
%


%% file: results/shapenet_ablation.tex

\begin{table}[!htbp]
\begin{small}
	\begin{tabular}{l|ccccc}
		\hline
		  		~				 &\hspace{-1mm} Chf. ($\downarrow$)& \hspace{-1mm}Normal & \hspace{-1mm} $F^{0.1}$  &  \hspace{-1mm}$F^{0.3}$  &\hspace{-1mm}  $F^{0.5}$\\ \Xhline{3\arrayrulewidth} 
\hspace{-1mm}		 \postprocessdasm{}			 &  \hspace{-1mm}  0.249     &  \hspace{-1mm}     0.673 			&  \hspace{-1mm}   28.5         &   \hspace{-1mm} 75.3    &  \hspace{-1mm} 88.1     \\  
\hspace{-1mm}                  \uniformdasm{} 		&\hspace{-1mm}  0.201            &  \hspace{-1mm}          0.699              &  \hspace{-1mm}   31.4   & \hspace{-1mm} 78.3&\hspace{-1mm} 91.3 \\
\hspace{-1mm}		 \adaptivedasm{} 			 &\hspace{-1mm}    \textbf{0.183}      &   \hspace{-1mm}    \textbf{0.727 }		& \hspace{-1mm}   \textbf{33.5 }  & \hspace{-1mm} \textbf{81.0}  &  \hspace{-1mm}    \textbf{{92.6}}    \\ \hline 
	\end{tabular}
\vspace{-2mm}
\caption{\small {\bf Ablation study on ShapeNet.}}
 \label{table:shapenet_ablation}

\end{small}
	
\end{table}

%% file: fig/results_shapenet.tex

\begin{figure*}
\centering
\includegraphics[height=8.75cm]{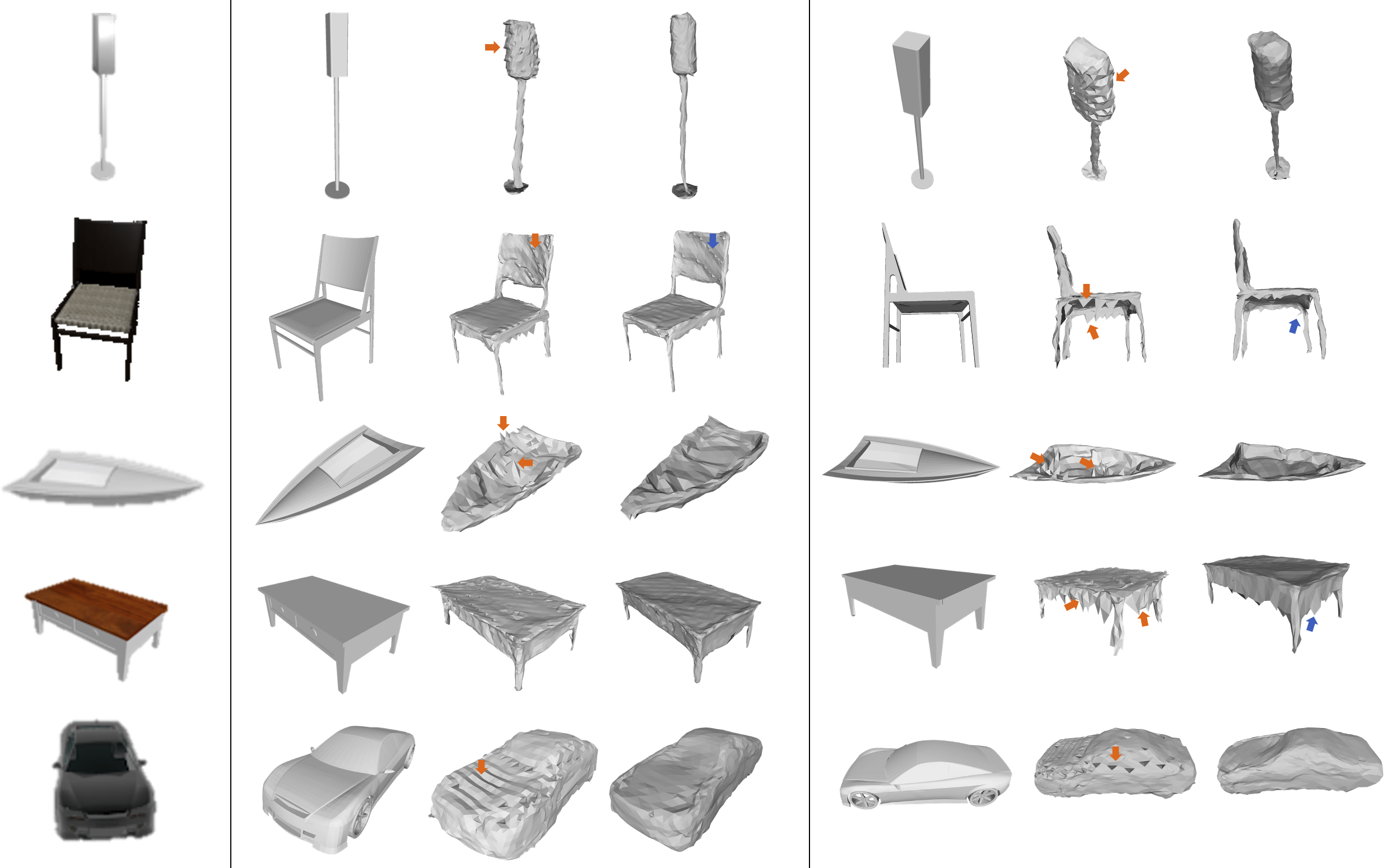}\\
\hspace{0cm}(a) \hspace{1.9cm}(b) \hspace{1.3cm}(c) \hspace{1.35cm}(d)  \hspace{1.9cm}(b) \hspace{1.3cm}(c) \hspace{1.3cm}(d) \hspace{1.0cm}\\
\hspace{4.4cm}View 1 \hspace{4.8cm} View 2 \hspace{1.8cm}
\vspace{-3mm}
\caption{\small {\bf ShapeNet Results.} (a) Input images (b) \meshrcnn{} results from two different viewpoints. The orange arrows highlight commonly seen \meshrcnn{} artifacts. (c) \dasm{} results in the same two views. The meshes are much smoother and most artifacts have disappeared, except for a few highlighted by blue arrows. }
\label{fig:shapenet_results}
\end{figure*}

%% file: fig/results_shapenet_lambda.tex

\begin{figure*}
\centering
\includegraphics[height=5.75cm]{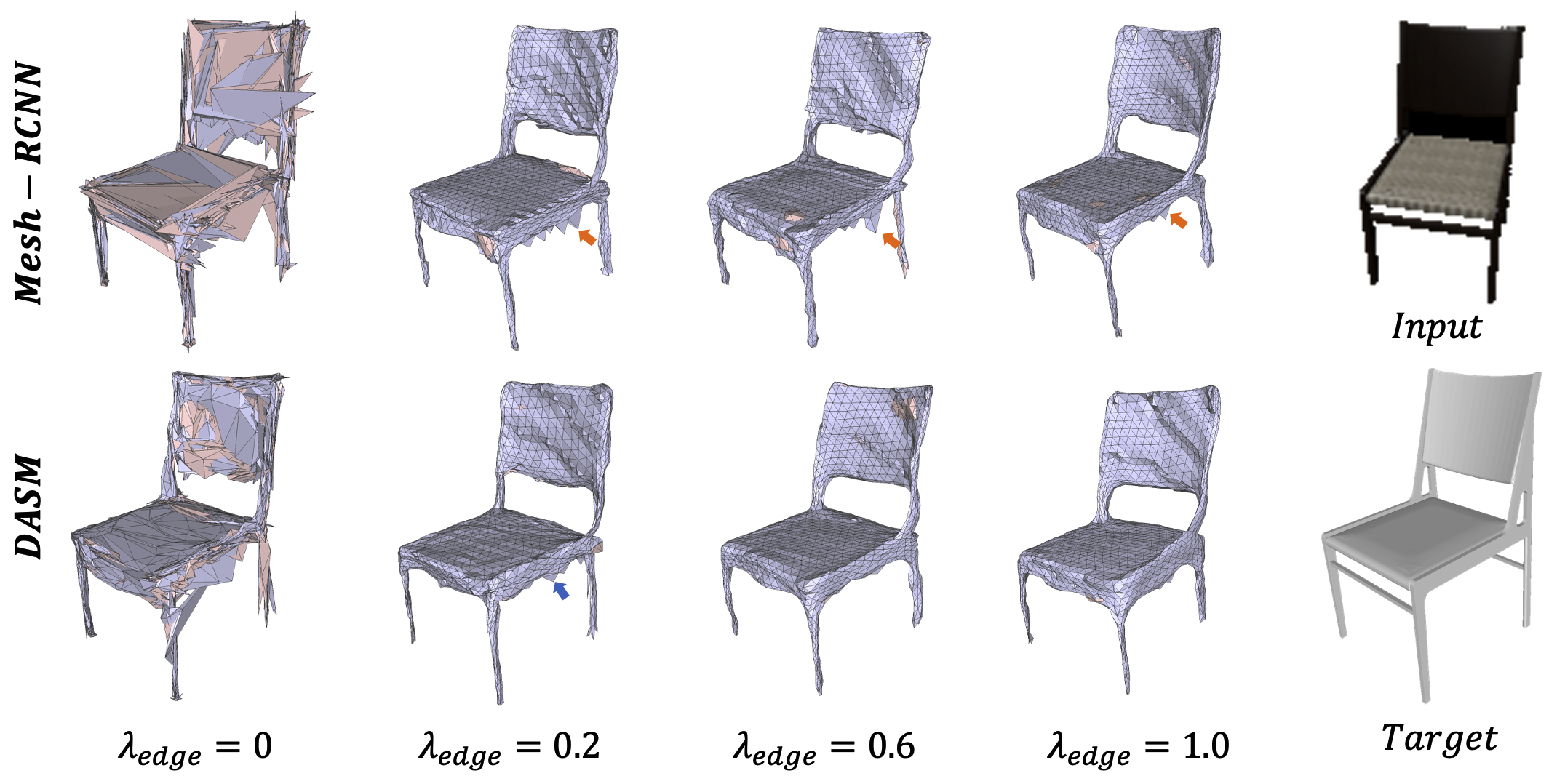}
\vspace{-3mm}
\caption{\small {\bf Influence of the regularization term.} As $\lambda_{reg}$ increases, the output of both methods becomes smoother but only \dasm{} completely eliminates the artifacts. Note that in the case $\lambda_{reg}=0$, the output of \meshrcnn{} is an extremely irregular mesh that nevertheless scores well on the Chamfer metric.}
\label{fig:shapenet_results_lambda}
\end{figure*}

%% file: fig/results_synapses.tex

\begin{figure*}
\centering
\includegraphics[height=7.25cm]{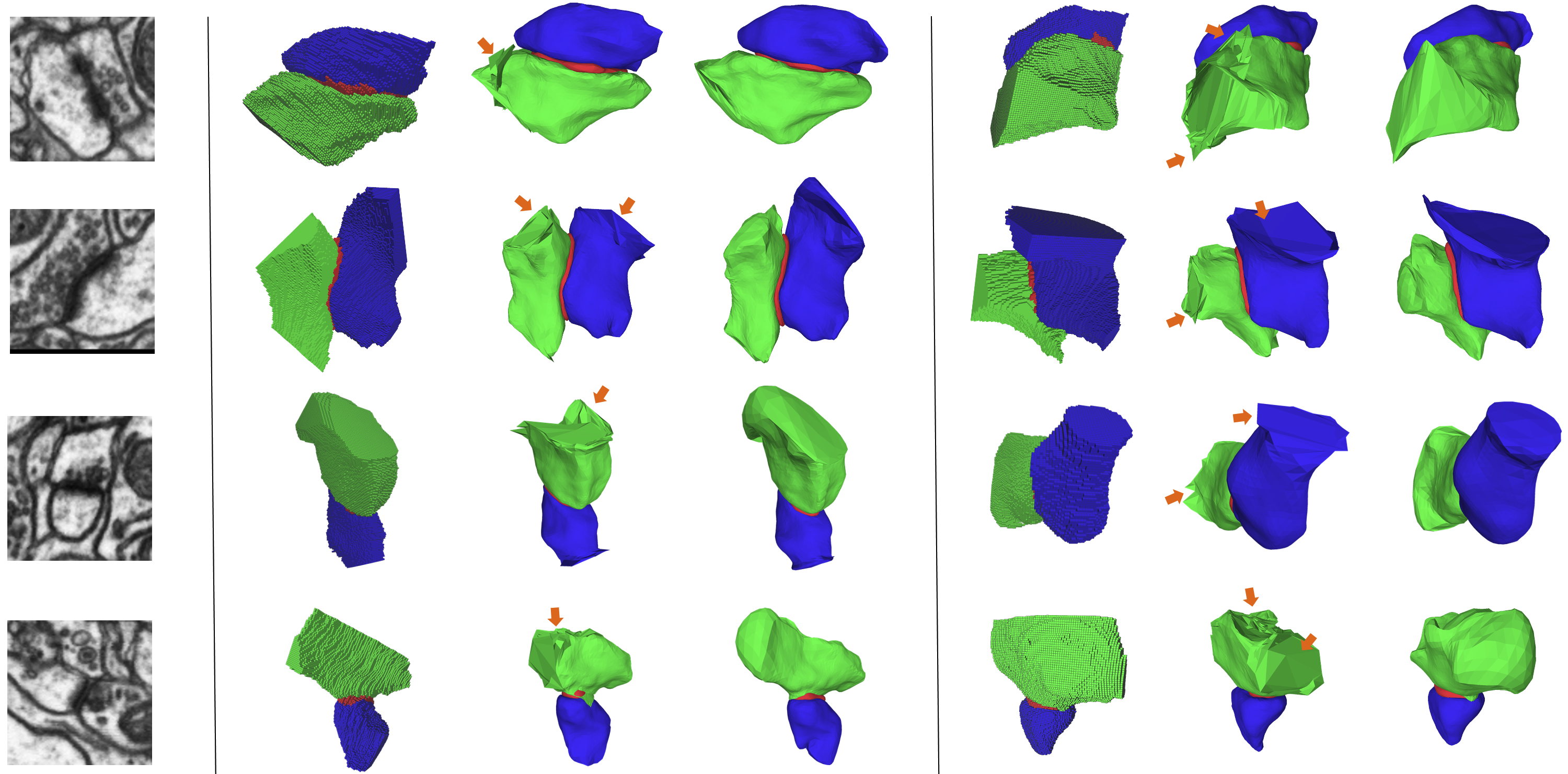}\\
\hspace{0cm}(a) \hspace{2.15cm}(b) \hspace{1.5cm}(c) \hspace{1.7cm}(d)  \hspace{1.75cm}(b) \hspace{1.45cm}(c) \hspace{1.55cm}(d) \hspace{3.35cm}\\

\hspace{3.4cm}View 1 \hspace{5.2cm} View 2 \hspace{0.6cm}
\vspace{-3mm}
\caption{\small{Modeling Synapses from Electron Microscopy Image Stacks} (a) Representative slice from the input volume. (b)  \voxmesh{} results seen from two different views. (c) \dasm{} results seen from the same two views. The pre-synaptic region, post-synaptic region, and synaptic cleft are shown in blue, green, and red, respectively. The \dasm{} results are much smoother and without artifacts, while also being more accurate.}
\label{fig:cortex_results}
\end{figure*}

%% file: results/crotex.tex

\begin{table}[htbp]
\begin{small}
\centering
\vspace{-3mm}

\begin{tabular}{|l|c|c|c|}
	\hline 
~															                   & Pre-Syn.             & Synapse       & Post-Syn.          \\ \hline
	\ternausnet{}     \cite{Iglovikov18}          & 73.5 $ \pm$1.3 & 64.4 $\pm$0.5 				& {78.4 $\pm$1.3}  \\ \hline
	\linknet{}     \cite{Shvets18}                  & 72.3 $ \pm$0.5 			& 63.2 $\pm$1.2 			& {78.2 $\pm$1.1}    \\ \hline
	\resnet{}     \cite{Kavur19b}                   & 70.3 $\pm$0.8 			& 63.3 $\pm$0.6 			& 76.2 $\pm$1.4     \\ \hline
	\resnetse{}  \cite{Kavur19b}                  & 71.3 $\pm$0.6  			& 63.6 $\pm$0.7 			& 76.3 $\pm$0.9  \\ \hline
	\vnet{}      \cite{Milletari16}                  & 64.3 $\pm$0.7  			& 65.2 $\pm$1.3 			& 74.1 $\pm$0.7     \\ \hline
	\unet{}     \cite{Cicek16}                     & {73.6 $\pm$1.3 } 	& \textbf{67.2 $\pm$0.8 }	& {78.2 $\pm$0.9}      \\ \Xhline{3\arrayrulewidth}
	\voxmesh{}$_{0.25 }$      & {77.3 $\pm$ 1.2}  	& 65.3 $\pm$1.2 			& {83.2 $\pm$1.6}         \\ \hline
	\voxmesh{}$_{0.025 }$   & {76.8 $\pm$ 0.9}  	& 65.4 $\pm$1.6 			& {81.2 $\pm$1.4}         \\ \Xhline{3\arrayrulewidth}
	 \adaptivedasm{}$_{0.25 }\!\!$         & {77.7 $\pm$0.8}  	& {65.3 $\pm$0.7 }			& {83.3 $\pm$0.8}  \\ \hline
	  \adaptivedasm{}$_{0.025 }\!\!\!$         & \textbf{79.4 $\pm$1.4}  	& {65.3 $\pm$0.9 }			& \textbf{85.5 $\pm$ 1.2}  \\ \hline
\end{tabular}
\vspace{-1.5mm}
\caption{\small {\bf Comparative results on CortexEM.} We use the IoU metric to compare volumetric segmentations.}
\label{table:main_results_cortex} 
\end{small}
\end{table}

%% file: results/crotex_chamfer.tex

\begin{table}[htbp]
\begin{small}
\centering
\vspace{-3mm}

\begin{tabular}{|l|c|c|c|}
	\hline 
~															                   & Pre-Syn.             & Synapse       & Post-Syn.           \\ \hline
\voxmesh{}$_{0.25 }$                              & 1.62 $\pm 0.4$	& 0.19 $\pm 0.2$ &   2.41  $\pm 0.7$    \\ \hline
\voxmesh{}$_{0.025 }$                         & 1.53 $\pm 0.6$	& \textbf{0.18 $\pm$ 0.4} &   2.35  $\pm 0.4$    \\ \hline
\adaptivedasm{}$_{0.25 }  \!\!$                         	  &  1.59 $\pm 0.5$ & 0.19  $\pm 0.3$& 2.39  $\pm 0.8$ \\ \hline
\adaptivedasm{}$_{0.025 } \!\!\!$            	  &  \textbf{1.47 $\pm$ 0.6} & 0.18  $\pm 0.5$& \textbf{2.31  $\pm$ 0.6} \\ \hline
\end{tabular}
\vspace{-1.5mm}
\caption{\small {\bf Comparative results on CortexEM.} We use the Chamfer distance ($\times 10^{-2}$)}
\label{table:main_results_chamfer} 
\end{small}
\end{table}

%% file: results/cortex_timing.tex

\begin{table}[htbp]

\begin{small}
\begin{center}
\begin{tabular}{|l|c|c|c|}
	\hline 
~															                   & Time (sec.)            & $|V|$               \\ \hline

\voxmesh{}                              & 1.06 $\pm 0.04$	& 1534 $\pm 4$     \\ \hline  
\adaptivedasm{}            	  &  1.47 $\pm$ 0.06 & 1535 $\pm 2$   \\ \hline
\end{tabular}
\end{center}
\vspace{-5mm}
\caption{\small {\bf Run times.} We report the time required to perform a forward and backward pass. $|V|$  is the number of  mesh vertices.}
\label{table:running_time} 
\end{small}
\end{table}

%% file: tex/5_conclussion.tex

\section{Conclusion}
 
 We  have developed an approach to incorporating Active Shape Models into layers that can be integrated seamlessly into  Graph Convolutional Networks to enforce sophisticated smoothness priors at an acceptable computational cost. By embedding the smoothing directly into the update equations, we can deliver smoother and more accurate meshes than methods that rely solely on a regularization loss term.

In future work, we will further improve our adaptive DASM scheme. Currently, it relies on a hand-designed metric to detect where to smooth and where not to. We will replace it by an auxiliary network that predicts  where smoothing is required, given the current state of the mesh and the input data.

%% file: tex/6_acknowledgment.tex
\section{Acknowledgment}
This work was supported in part by the Swiss National Science Foundation. 

%% file: tex/7_appendix.tex
 
\section{Appendix} 
\label{sec:appendix}

\subsection{Active Surface Model: Continuous Formulation}

Our objective is to minimize the total energy $\bE$ in Eq.~\ref{eq:minimize1}. There is no analytical solution for the global minimum  of $\bE$. But, as mentioned in Section~\ref{sec:asm}, any local minimum must satisfy the associated Euler-Lagrange equation given in Eq.~\ref{eq:euler}. To find a surface that does this, surface evolution is used by introducing a time $t$ parameter into \ref{eq:euler} and writing 
\begin{equation}\label{eq:evolution}
\frac{\partial v(s,r,t; \Phi)}{\partial t} + L(v(s,r,t; \Phi)) = F(v(s,r,t; \Phi)) \; ,
\end{equation}
where $L(v(s,r,t; \Phi))$ is the R.H.S of Eq.~\ref{eq:euler}.

Solving \ref{eq:evolution}, requires specifying an initial surface. Earlier approaches \cite{Cohen93,Kass88} used a manual initialization, whereas in~\cite{Marcos18,Cheng19} another model is used to predict the initial curve. To ensure the reached local minima corresponds to the desired curve, these approaches require the initialization to be close to the target shape. In DASM, we rely instead on the graph-convolution layers to provide a good initialization. 

\subsection{Active Surface Model: Discrete Formulation}

In the continuous formulation of Section~\ref{sec:asm}, computing the solution to Eq.~\ref{eq:euler} requires computing the derivatives of order 2 and 4 for the mapping $v$ of Eq.~\ref{eq:surface}. To compute them in practice, we discretize the surface and use finite difference equations to estimate the derivatives. Given a small value of $\delta s$, finite-difference approximations for the derivatives w.r.t $s$ can be written as, 
\begin{small}
	\begin{align*}
	&\frac{\partial v}{\partial s} \approx  \frac{1}{\delta s} [v(s + \delta s, r) - v(s, r) ] \; ,\\ 
	&\frac{\partial^2 v}{\partial s^2} \approx  \frac{1}{\delta s^2} [v(s + \delta s, r) - 2v(s, r) + v(s - \delta s, r)]\; ,\\ 
	&\frac{\partial^3 v}{\partial s^3} \approx \frac{1}{\delta s^3} [v(s+ 2\delta s, r) - 3 v(s + \delta s, r)  \; \\ 
	& ~~~~~~~~~~~~~~~~~~~~~~ ~~+ 3v(s, r) - v(s - \delta s, r)] \; ,    \nonumber \\ 
	&\frac{\partial^4 v}{\partial s^4} \approx \frac{1}{\delta s^4} [v(s+ 2 \delta s, r) - 4 v(s + \delta s, r)  \; \\ 
    & ~~~~~~~~~~~~~~~~~~~~~~ ~~+ 6 v(s, r) - 4 v(s - \delta s, r) + v(s - 2 \delta s, r)]  \; ,    \nonumber  
	\end{align*}
\end{small}Similarly, we can write finite difference equations w.r.t $r$ as well. 
%
\input{fig/finite_differences}
Now to compute these approximations, we need to compute $v(s+\delta s)$ and other similar terms. Let us therefore take $(s,r)$ be the 2D coordinates that $v$ maps to the coordinates of a specific vertex. In an irregular grid, $(s,r+\delta r)$, $(s+\delta s,r)$, or any of $s,r$ coordinates that appear in the derivative computations will in general {\it not} be be mapped to another vertex for any choice of $\delta s, \delta r$. Fig.~\ref{fig:finite_differences} illustrates their actual positions depending on the number of neighbors the vertex has.


We can nevertheless compute the 3D coordinates they map to as follows. Let us first consider the 3D point $v(s+\delta s,r)$ that $(s+\delta s,r)$ gets mapped to  and it is depicted by orange circle in Fig.~\ref{fig:surface}. For $\delta s$ small enough, it belongs to a facet of which  $v(s,r)$ is a vertex and let   $v(s_1,r_1)$ and  $v(s_2,r_2)$ be the other two. We can compute the barycentric coordinates $\lambda$, $\lambda_1$, and $\lambda_2$ of $v(s+\delta s,r)$ in that facet by solving
\begin{align}\label{eq:barycentric}
 \begin{bmatrix}
s + \delta s\\
r\\
1
\end{bmatrix} =& 
 \begin{bmatrix}
s & s_{1} & s_{2} \\
r & r_{1} & r_{2}\\
1 & 1 & 1
\end{bmatrix} 
 \begin{bmatrix}
\lambda\\
\lambda_{1}\\
\lambda_{2}\\
\end{bmatrix} \; .
\end{align}
Given these barycentric coordinates, we can now estimate $ v(s+\delta s, r)$ as 
\begin{equation}\label{eq:weightedSum} 
\lambda*v(s,r) + \lambda_1*v(s_1,r_1) + \lambda_2*v(s_2,r_2) \; ,
\end{equation}
which allows us to estimate $\frac{\partial v}{\partial s}$ according to the above finite-difference equations. For this approximation to be valid, we pick $\delta s$ such that all terms in finite-difference expressions lie within the 1-ring neighborhood of $v(s,r)$. We can repeat the process for all the other expressions involving $\delta s$ in these equations and, hence, compute all required derivatives.  Regular square and hexagonal grids are special cases in which these computations can be simplified. 

\subsection{Matrix Inversion using Neumann Series}

We are approximating the inverse of $(\bA + \alpha\bI)^{-1}$ using the Neumann series given in Eq.~\ref{eq:inverse}. In Fig~\ref{fig:neuman}, we plot both RMSE in estimating the inverse and the time it takes to perform the estimation as a function of $K$. Given the trade off between running time and accuracy, we pick $K=4$ for the estimation.

\input{fig/neuman}

\subsection{Quantitatively Measuring Mesh Regularization with Consistency Metrics}

All the metrics used in Sec.~\ref{sec:experiments} evaluate the accuracy of the meshes.  We use them because they are the standard metrics used in the literature. But to get a better understanding of the quality of the meshes, we provide two more metrics;  mean edge length and mean surface Laplacian. We observe that around abnormalities such as those highlighted by orange arrows in Fig.~\ref{fig:shapenet_results}, \ref{fig:cortex_results}, the edge lengths and surface Laplacians tend to increase significantly. This increases mean edge length and mean surface Laplacian and its effect can be seen in Table~\ref{table:shapenet_results_diff_lambda_new}.

\input{results/shapenet_diff_lambda_new.tex}

%% file: fig/finite_differences.tex

\begin{figure}
\centering
\includegraphics[height=8.0cm]{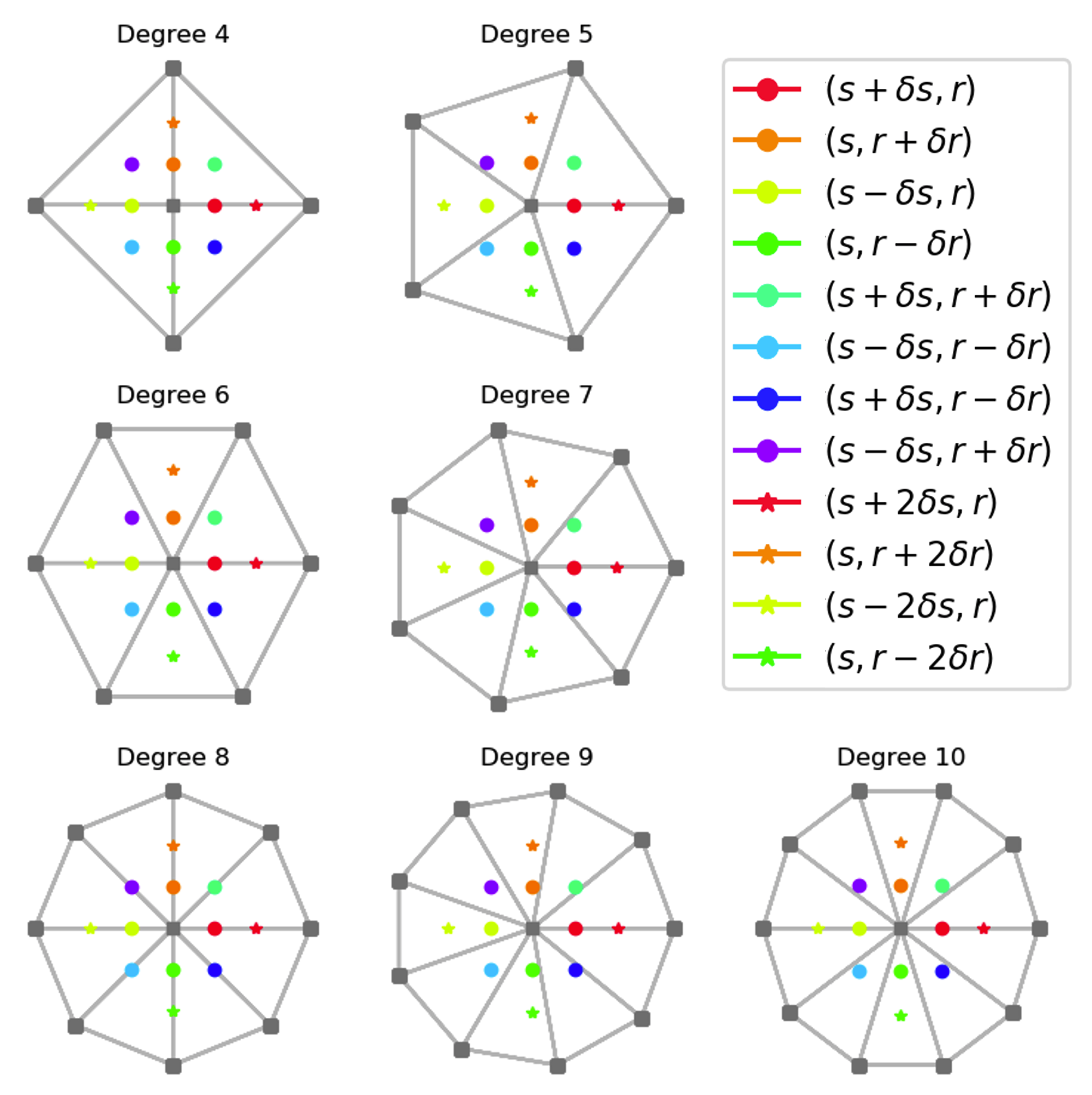}
\caption{\small{ {\bf Finite Differences.} Relative positions of $(s + k_1\delta, r + k_2\delta)$ terms w.r.t $(s,r)$ which is at the center and its 1-ring neighbors from degree 4 to 10.}}
\label{fig:finite_differences}
\end{figure}

%% file: fig/neuman.tex

\begin{figure}[htbp]
\centering
\includegraphics[height=6.25cm]{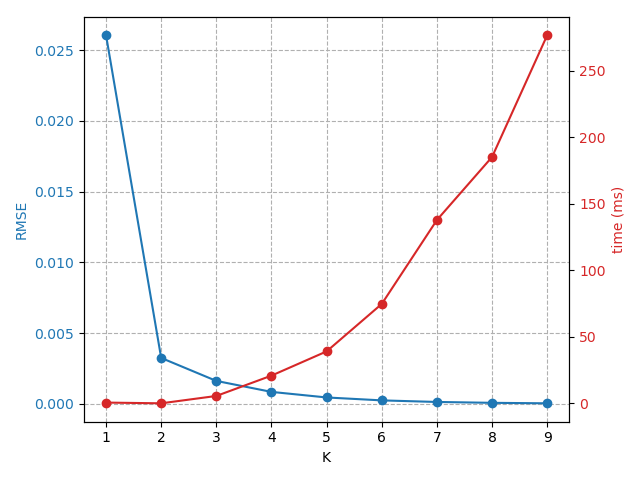}
\caption{\small {\bf Neumann Series Approximation.} RMSE between the approximated inverse and the true one in blue and computation time in red as function of $K$. $K=4$ gives an acceptable trade-off between the two.}
\label{fig:neuman}
\end{figure}

%% file: results/shapenet_diff_lambda_new.tex
  
\begin{table}[!htbp]

\begin{small}
	\begin{tabular}{l|l|cccc}
		\hline
	$\lambda_{e.}$	 			&  				\hspace{-\x mm}		& \hspace{-\x mm} Chf. ($\downarrow$)  	 				& \hspace{-\x mm} Edg. Length		 	 & \hspace{-\x mm} Surf. Lap. 			\\ \Xhline{3\arrayrulewidth}
	\multirow{2}{*}{1.0}    	& \hspace{-\y mm}\meshrcnn{}  	    & \hspace{-\x mm}	0.232                       		& \hspace{-\x mm}0.023 $\pm$ 0.011   & \hspace{-\x mm}0.033 $\pm$ 0.031		 \\
								& \hspace{-\y mm}\adaptivedasm{}	& \hspace{-\x mm}\textbf{0.231}                     	& \hspace{-\x mm}0.022 $\pm$ 0.009	 & \hspace{-\x mm}0.023 $\pm$ 0.019   	 \\ \hline
 
	\multirow{2}{*}{0.6}    	& \hspace{-\y mm}\meshrcnn{}        & \hspace{-\x mm}   0.212     							&  \hspace{-\x mm}0.024 $\pm$ 0.015   & \hspace{-\x mm}0.045 $\pm$ 0.045 	\\
								& \hspace{-\y mm}\adaptivedasm{} 	& \hspace{-\x mm}   \textbf{0.206}      				&  \hspace{-\x mm}0.023 $\pm$ 0.011   & \hspace{-\x mm}0.029 $\pm$ 0.020	\\ \hline

	\multirow{2}{*}{0.2}    	& \hspace{-\y mm}\meshrcnn{} 	    & \hspace{-\x mm}   0.189      		    				&  \hspace{-\x mm}0.028 $\pm$ 0.021  & \hspace{-\x mm} 0.066 $\pm$ 0.075	\\
								& \hspace{-\y mm}\adaptivedasm{} 	& \hspace{-\x mm}   \textbf{0.183}      				&  \hspace{-\x mm}0.025 $\pm$ 0.015  & \hspace{-\x mm} 0.037 $\pm$ 0.049	\\ \hline
	 
	\multirow{2}{*}{0.0}    	& \hspace{-\y mm}\meshrcnn{} 	    & \hspace{-\x mm}   \textbf{0.144}     					&  \hspace{-\x mm}0.141 $\pm$ 0.142  & \hspace{-\x mm}0.708 $\pm$ 0.671		\\
								& \hspace{-\y mm}\adaptivedasm{} 	& \hspace{-\x mm}    0.167              				&  \hspace{-\x mm}0.070 $\pm$ 0.072  & \hspace{-\x mm}0.254 $\pm$ 0.276		 \\ \hline
	\end{tabular}
\vspace{-2mm}
\caption{\small {\bf Results on ShapeNet} as a function of $\lambda_{edge}$. Note that this is an extension of Table~\ref{table:shapenet_results_diff_lambda}. }
\label{table:shapenet_results_diff_lambda_new}

\end{small}
\end{table}

%
%
%


%% file: top.bbl
\begin{thebibliography}{10}\itemsep=-1pt

\bibitem{Acuna19}
D. Acuna, A. Kar, and S. Fidler.
\newblock {Devil is in the Edges: Learning Semantic Boundaries from Noisy
  Annotations}.
\newblock In {\em Conference on Computer Vision and Pattern Recognition}, 2019.

\bibitem{Kavur19b}
A.~Kavur amd M.~Selver.
\newblock {CHAOS Challenge - Combined (CT-MR) Healthy Abdominal Organ
  Segmentation.}
\newblock 2020.

\bibitem{Chang15}
A. Chang, T. Funkhouser, L. G., P. Hanrahan, Q. Huang, Z. Li, S. Savarese, M.
  Savva, S. Song, H. Su, J. Xiao, L. Yi, and F. Yu.
\newblock {Shapenet: An Information-Rich 3D Model Repository}.
\newblock In {\em arXiv Preprint}, 2015.

\bibitem{Chen19c}
Z. Chen and H. Zhang.
\newblock {Learning Implicit Fields for Generative Shape Modeling}.
\newblock In {\em Conference on Computer Vision and Pattern Recognition}, 2019.

\bibitem{Cheng19}
D. Cheng, R. Liao, S. Fidler, and R. Urtasun.
\newblock {DARNet: Deep Active Ray Network for Building Segmentation}.
\newblock In {\em Conference on Computer Vision and Pattern Recognition}, 2019.

\bibitem{Chibane20}
J. Chibane, T. Alldieck, and G. Pons{-}Moll.
\newblock {Implicit Functions in Feature Space for 3D Shape Reconstruction and
  Completion}.
\newblock In {\em Conference on Computer Vision and Pattern Recognition}, 2020.

\bibitem{Choy16}
C. Choy, D. Xu, J. Gwak, K. Chen, and S. Savarese.
\newblock {3D-R2N2: A unified approach for single and multi-view 3d object
  reconstruction}.
\newblock In {\em European Conference on Computer Vision}, 2016.

\bibitem{Cicek16}
{\"O}. {\c{C}}i{\c{c}}ek, A. Abdulkadir, S. Lienkamp, T. Brox, and O.
  Ronneberger.
\newblock {3D U-Net: Learning Dense Volumetric Segmentation from Sparse
  Annotation}.
\newblock In {\em Conference on Medical Image Computing and Computer Assisted
  Intervention}, pages 424--432, 2016.

\bibitem{Cohen93}
L.D. Cohen and I. Cohen.
\newblock {Finite-Element Methods for Active Contour Models and Balloons for 2D
  and 3D Images}.
\newblock {\em IEEE Transactions on Pattern Analysis and Machine Intelligence},
  15(11):1131--1147, November 1993.

\bibitem{Dong18b}
S. Dong and H. Zhang.
\newblock {A Combined Fully Convolutional Networks and Deformable Model for
  Automatic Left Ventricle Segmentation Based on 3D Echocardiography}.
\newblock In {\em BioMed Research International}, 2018.

\bibitem{Duchi11}
J. Duchi, E. Hazan, and Y. Singer.
\newblock {Adaptive Subgradient Methods for Online Learning and Stochastic
  Optimization}.
\newblock In {\em Journal of Machine Learning Research}, 2011.

\bibitem{Fua96f}
P. Fua.
\newblock {Model-Based Optimization: Accurate and Consistent Site Modeling}.
\newblock In {\em International Society for Photogrammetry and Remote Sensing},
  July 1996.

\bibitem{Fua95c}
P. Fua and Y.~G. Leclerc.
\newblock {Object-Centered Surface Reconstruction: Combining Multi-Image Stereo
  and Shading}.
\newblock {\em International Journal of Computer Vision}, 16:35--56, September
  1995.

\bibitem{Gkioxari19}
G. Gkioxari, J. Malik, and J. Johnson.
\newblock {Mesh R-CNN}.
\newblock In {\em International Conference on Computer Vision}, 2019.

\bibitem{Hatamizadeh20}
A. Hatamizadeh, D. Sengupta, and D. Terzopoulos.
\newblock {End-To-End Trainable Deep Active Contour Models for Automated Image
  Segmentation: Delineating Buildings in Aerial Imagery}.
\newblock In {\em arXiv Preprint}, 2020.

\bibitem{He08a}
L. He, Z. Peng, B. E., X. Wang, C.~Y. Han, K.~L. Weiss, and W.~G. Wee.
\newblock {A Comparative Study of Deformable Contour Methods on Medical Image
  Segmentation}.
\newblock {\em Image and Vision Computing}, 26(2):141--163, 2008.

\bibitem{Iglovikov18}
V. Iglovikov and A. Shvets.
\newblock {Ternausnet: U-Net with VGG11 Encoder Pre-Trained on Imagenet for
  Image Segmentation}.
\newblock In {\em arXiv Preprint}, 2018.

\bibitem{Jorstad14a}
A. Jorstad, B. Nigro, C. Cali, M. Wawrzyniak, P. Fua, and G.W. Knott.
\newblock {Neuromorph: A Toolset for the Morphometric Analysis and
  Visualization of 3D Models Derived from Electron Microscopy Image Stacks}.
\newblock {\em Neuroinformatics}, 13(1):83--92, 2014.

\bibitem{Kass88}
M. Kass, A. Witkin, and D. Terzopoulos.
\newblock {Snakes: Active Contour Models}.
\newblock {\em International Journal of Computer Vision}, 1(4):321--331, 1988.

\bibitem{Kingma14a}
D.~P. Kingma and J. Ba.
\newblock {Adam: {A} Method for Stochastic Optimization}.
\newblock In {\em International Conference on Learning Representations}, 2015.

\bibitem{Lengagne97}
R. Lengagne, P. Fua, and O. Monga.
\newblock {Using Differential Constraints to Reconstruct Complex Surfaces from
  Stereo}.
\newblock In {\em Conference on Computer Vision and Pattern Recognition}, 1997.

\bibitem{Lengagne00}
R. Lengagne, P. Fua, and O. Monga.
\newblock {3D Stereo Reconstruction of Human Faces Driven by Differential
  Constraints}.
\newblock {\em Image and Vision Computing}, 18(4):337--343, March 2000.

\bibitem{Leventon00}
M.~E. Leventon, W.~E. Grimson, and O. Faugeras.
\newblock {Statistical Shape Influence in Geodesic Active Contours}.
\newblock In {\em Conference on Computer Vision and Pattern Recognition}, pages
  316--323, 2000.

\bibitem{Liang20}
J. Liang, N. Homayounfar, W. Ma, Y. Xiong, R. Hu, and R. Urtasun.
\newblock {Polytransform: Deep Polygon Transformer for Instance Segmentation}.
\newblock In {\em Conference on Computer Vision and Pattern Recognition}, 2020.

\bibitem{Ling19}
H. Ling, J. Gao, A. Kar, W. Chen, and S. Fidler.
\newblock {Fast Interactive Object Annotation with Curve-Gcn}.
\newblock In {\em Conference on Computer Vision and Pattern Recognition}, pages
  5257--5266, 2019.

\bibitem{Lorensen87}
W.E. Lorensen and H.E. Cline.
\newblock {Marching Cubes: {A} High Resolution 3{D} Surface Construction
  Algorithm}.
\newblock In {\em ACM SIGGRAPH}, pages 163--169, 1987.

\bibitem{Marcos18}
D. Marcos, D. Tuia, B. Kellenbergerg, and R. Urtasun.
\newblock {Learning Deep Structured Active Contours End-To-End}.
\newblock In {\em Conference on Computer Vision and Pattern Recognition}, 2018.

\bibitem{McInerney95a}
T. Mcinerney and D. Terzopoulos.
\newblock {A Dynamic Finite Element Surface Model for Segmentation and Tracking
  in Multidimensional Medical Images with Application to Cardiac 4D Image
  Analysis}.
\newblock {\em Computerized Medical Imaging and Graphics}, 19(1):69--83, 1995.

\bibitem{Mescheder19}
L. Mescheder, M. Oechsle, M. Niemeyer, S. Nowozin, and A. Geiger.
\newblock {Occupancy Networks: Learning 3D Reconstruction in Function Space}.
\newblock In {\em Conference on Computer Vision and Pattern Recognition}, pages
  4460--4470, 2019.

\bibitem{Milletari16}
F. Milletari, N. Navab, and S.-A. Ahmadi.
\newblock {V-Net: Fully Convolutional Neural Networks for Volumetric Medical
  Image Segmentation}.
\newblock In {\em arXiv Preprint}, June 2016.

\bibitem{Newman06}
T.S. Newman and H. Yi.
\newblock {A Survey of the Marching Cubes Algorithm}.
\newblock {\em Computers \& Graphics}, 30(5):854--879, 2006.

\bibitem{Pan19}
J. Pan and K. Jia.
\newblock {Deep Mesh Reconstruction from Single RGB Images via Topology
  Modification Networks}.
\newblock In {\em International Conference on Computer Vision}, 2019.

\bibitem{Park20a}
J.~J. Park, P. Florence, J. Straub, R.~A. Newcombe, and S. Lovegrove.
\newblock {Deepsdf: Learning Continuous Signed Distance Functions for Shape
  Representation}.
\newblock In {\em Conference on Computer Vision and Pattern Recognition}, 2019.

\bibitem{Peng20}
S. Peng, W. Jiang, H. Pi, X. Li, H. Bao, and X. Zhou.
\newblock {Deep Snake for Real-Time Instance Segmentation}.
\newblock In {\em Conference on Computer Vision and Pattern Recognition}, 2020.

\bibitem{Prevost13}
R. Prevost, R. Cuingnet, B. Mory, D. {L.D. C.}, and R. Ardon.
\newblock {Incorporating Shape Variability in Image Segmentation via Implicit
  Template Deformation}.
\newblock {\em Conference on Medical Image Computing and Computer Assisted
  Intervention}, pages 82--89, 2013.

\bibitem{Remelli20b}
E. Remelli, A. Lukoianov, S. Richter, B. Guillard, T. Bagautdinov, P. Baque,
  and P. Fua.
\newblock {Meshsdf: Differentiable Iso-Surface Extraction}.
\newblock In {\em Advances in Neural Information Processing Systems}, 2020.

\bibitem{Shvets18}
A. Shvets, A. Rakhlin, A. Kalinin, and V. Iglovikov.
\newblock {Automatic Instrument Segmentation in Robot-Assisted Surgery Using
  Deep Learning}.
\newblock In {\em arXiv Preprint}, 2018.

\bibitem{Terzopoulos87}
D. Terzopoulos, A. Witkin, and M. Kass.
\newblock {Symmetry-Seeking Models and 3D Object Reconstruction}.
\newblock {\em International Journal of Computer Vision}, 1:211--221, 1987.

\bibitem{Terzopoulos88}
D. Terzopoulos, A. Witkin, and M. Kass.
\newblock {Constraints on Deformable Models: Recovering 3D Shape and Nonrigid
  Motion}.
\newblock {\em Artificial Intelligence}, 36(1):91--123, 1988.

\bibitem{Wang18e}
N. Wang, Y. Zhang, Z. Li, Y. Fu, W. Liu, and Y. Jiang.
\newblock {Pixel2mesh: Generating 3D Mesh Models from Single RGB Images}.
\newblock In {\em European Conference on Computer Vision}, 2018.

\bibitem{Wen19a}
C. Wen, Y. Zhang, Z. Li, and Y. Fu.
\newblock {Pixel2mesh++: Multi-View 3D Mesh Generation via Deformation}.
\newblock In {\em International Conference on Computer Vision}, 2019.

\bibitem{Wickramasinghe20}
U. Wickramasinghe, E. Remelli, G. Knott, and P. Fua.
\newblock Voxel2mesh: 3d mesh model generation from volumetric data.
\newblock In {\em Conference on Medical Image Computing and Computer Assisted
  Intervention}, 2020.

\bibitem{Xu19b}
Q. Xu, W. Wang, D. Ceylan, R. Mech, and U. Neumann.
\newblock {DISN: Deep Implicit Surface Network for High-Quality Single-View 3D
  Reconstruction}.
\newblock In {\em Advances in Neural Information Processing Systems}, 2019.

\end{thebibliography}
